\documentclass[pdftex,sn-mathphys-num,14pt]{sn-jnl}

\usepackage{graphicx}
\usepackage{booktabs}
\usepackage{amsmath,amssymb}
\usepackage{hyperref}
\usepackage{cleveref}

\usepackage{graphicx}%
\usepackage{multirow}%
\usepackage{amsmath,amssymb,amsfonts}%
\usepackage{amsthm}%
\usepackage{mathrsfs}%
\usepackage[title]{appendix}%
\usepackage{xcolor}%
\usepackage{textcomp}%
\usepackage{manyfoot}%
\usepackage{booktabs}%
\usepackage{algorithm}%
\usepackage{algorithmicx}%
\usepackage{algpseudocode}%
\usepackage{listings}%
\usepackage{adjustbox}

\theoremstyle{thmstyleone}

\usepackage[most]{tcolorbox}

\begin{document}
	
	\title[]{\large \textbf{Enhancing Deep Learning Model Performance through Data Generation Strategies: Classification of Olive Tree Diseases}}

\author*[1]{\fnm{Hicham} \sur{El Akhal}}
\email{hi.elakhal@edu.umi.ac.ma}

\author[2]{\fnm{Abderrahim} \sur{Waga}}
\email{a.waga@ueuromed.org}

\author[1]{\fnm{Aissa} \sur{Ben Yahya}}
\email{ai.benyahya@edu.umi.ac.ma}

\author[1]{\fnm{Abdelbaki} \sur{El Belrhiti El Alaoui}}
\email{a.elbelrhitielalaoui@umi.ac.ma}

\affil[1]{%
  \orgdiv{Computer Science and Applications Laboratory},
  \orgname{Faculty of Sciences, Moulay Ismail University of Meknes},
  \orgaddress{
    \street{PB 11201 Zitoune},
    \city{Meknes},
    \postcode{50000},
    \country{Morocco}
  }
}

\affil[2]{%
  \orgname{Euromed University of Fez},
  \orgaddress{
    \city{Fez},
    \country{Morocco}
  }
}


\abstract{
		Deep learning models are highly effective in a wide range of domains, particularly in plant disease diagnosis. However, their performance is often limited by challenges such as insufficient data and class imbalance. In our work, we propose a comprehensive data generation strategy to overcome these limitations. First, we apply various data generation techniques, including traditional augmentation methods, CutMix, MixUp, and synthetic image generation using Deep Convolutional Generative Adversarial Networks (DCGANs), to create new data samples.
Second, we exploit this generated data in three scenarios: (S1) balancing class distribution in the original dataset, (S2) increasing the number of samples per class, and (S3) expanding the dataset while maintaining balanced classes. Third, we evaluate the impact of these data distributions (S1, S2, S3) on model performance using the EfficientNetV2B0 architecture with pre-trained and fine-tuned weights. The experimental results show a substantial improvement in classification accuracy, emphasizing the critical role of data diversity and class balance. These findings highlight the effectiveness and adaptability of the proposed strategy of data generation and distribution for enhancing dataset quality, offering promising applications in plant disease detection and precision agriculture.
	    }




\maketitle

\section{Introduction}

Plant diseases are a serious challenge to agricultural production, leading to significant crop losses and threatening global food security \cite{caparas2021increasing}. Traditional detection techniques, which mostly rely on visual inspection and manual evaluation, are frequently laborious, subjective, and prone to human error. These issues are made worse in places where it is hard to reach agricultural specialists, as a delayed diagnosis can lead to a large decrease in productivity.

In the field of agriculture, artificial intelligence (AI) has emerged as a revolutionary technology, particularly in the areas of plant disease detection and surveillance. AI enables the early and precise identification of these diseases by leveraging extensive data, including sensor and image information \cite{liu2020artificial,attri2023machine}. Because it can automatically learn layered characteristics from raw data, deep learning, a subfield of artificial intelligence, plays a crucial role in this situation by reducing the need for intricate preprocessing or manually created features. Using supervised and unsupervised learning techniques, advanced analysis models have been effectively used for tasks like finding sick areas, classifying plant diseases, and determining their severity \cite{hasan2020review}. These models have proved highly effective in detecting disease in a wide range of crops, including apple \cite{9118928,sym10010011}, citrus \cite{9481921, CETINER2022104687}, grapevine \cite{gr}, olive \cite{FAZARI2021106252,alshammari2022optimal, ELAKHAL2023102276,el2025positive}, and tomato \cite{ANANDHAKRISHNAN2022100793,karthik2020attention,kaur2022approach}.

Despite these achievements, the lack of diverse and well-balanced datasets remains a significant challenge. The effectiveness of models can be negatively impacted by overfitting and limited generalization, which may arise from an imbalanced or biased data distribution. However, techniques such as federated learning (FL), which combine decentralized data from multiple sources, seek to address this issue \cite{li2020review,mamba2023image}. This research adopts a different approach, focusing on the generation of new data.

In this study, we generate enriched training data using a centralized single-server infrastructure. Our approach incorporates traditional data generation methods. \cite{enkvetchakul2022effective}, as well as more advanced strategies such as CutMix \cite{pan2024enhanced}, Mixup \cite{lewy2023overview}, and deep convolutional generative adversarial networks (DCGANs) \cite{deshpande2022tomato,chen2023grape}. These methods aim to improve the performance of convolutional neural networks (CNNs) in classifying olive tree diseases.

The main goal of this research is to create a solid framework for data generation that will improve datasets used to categorize diseases of olive trees. We focus on three specific scenarios: \textbf{(S1)} balancing the classes in the original dataset, \textbf{(S2)} increasing the number of samples per class, and \textbf{(S3)}expanding the dataset for all classes while maintaining a balanced class distribution.

Through a generative diversity strategy addressing the limitation of having small datasets, the proposed model endeavors to enhance the precision and usability of models in diagnosing plant disease. Unlike previous research that largely relies on primitive augmentation techniques, this paper presents a sophisticated pipeline that integrates traditional transformations with generative models, whose capacity can greatly improve the robustness of the models and facilitate their utilization in various applications of agriculture, especially in sparse data scenarios. This is especially significant in low-access-data situations.

The subsequent sections provide a detailed discussion in \autoref{sec5}, present the results in \autoref{sec4}, outline the experimental methodology in \autoref{sec3}, and review the relevant literature in \autoref{sec2}. \autoref{sec6}, the study's conclusion, offers possible avenues for further investigation.

\section{Literature Review}\label{sec2}

\subsection{Data Generation and Augmentation Strategies}

This section examines recent developments in data creation and augmentation techniques intended to tackle issues like class imbalance and data scarcity in image-based classification problems. Their use in plant disease detection is highlighted in particular, since these methods have been crucial in enhancing models' resilience, generality, and accuracy.

To improve plant disease classification, \cite{tabbakh2023deep} suggested TLMViT, a hybrid model that combines Vision Transformers (ViT) and transfer learning. By incorporating efficient data augmentation techniques, their model successfully mitigated overfitting and achieved 99.86\% accuracy on a wheat disease dataset.

\cite{mhala2025enhancing} used data augmentation, L2 regularization, and transfer learning to address class imbalance in potato disease datasets. Their final result was 81.31\%, with a 4.17\% increase in classification accuracy using DenseNet201.

\cite{singh2023zero} investigated domain generalization using a framework for zero-shot transfer learning. Even in cases where domain classes did not coincide, their approach showed strong generalization capabilities by fusing data augmentation with synthetic data synthesis.

\cite{sathya2025enhancing} created artificial images of both healthy and ill leaves using conditional GANs. An attention-augmented ResNet combined with Faster R-CNN was trained using these, yielding a classification accuracy of 98.78\%.

\cite{wu2023classification} added attention mechanisms to a ConvNeXt-based architecture by integrating random masking into their data augmentation pipeline. Using this method, the average accuracy on a soybean disease classification task was 85.42\%.

\cite{karthikeyan2023deep} created DTLDN-CBIRA, a content-based image retrieval system based on DenseNet. With 81.90\% recall and an F1-score of 89.90\%, their model reached 100\% accuracy by augmenting limited training data with rotation and mirroring.

\cite{haikal2024comprehensive} presented a framework for composite augmentation that combines FMix, MixUp, Cutout, and CutMix. When used to classify rice leaf diseases, the method produced accuracy rates of 99.75\% with MobileNetV3 and 91.51\% with ResNet34.

\cite{chen2024improved} CNN and ViT modules were integrated to improve the FasterViT architecture. Their method, which combined MixUp, Cutout, Triplet Attention, and Adaptive Average Pooling, showed better results on PlantVillage and small-scale datasets.

A summary of these approaches is presented in \autoref{r}, which highlights the main data generation and augmentation strategies for classification tasks.

\begin{table}[!ht]
	\centering
	\caption{Summary of Related Work on Data Generation and Augmentation Strategies for Classification}
	\label{r}
	\begin{tabular}{@{}p{2.5cm}p{4cm}p{3cm}p{4cm}@{}}
		\hline
		\textbf{Study} & \textbf{Methodology} & \textbf{Dataset} & \textbf{Performance} \\ \hline
		\cite{tabbakh2023deep} & Hybrid model (TLMViT) with transfer learning & Wheat disease dataset & 99.86\% accuracy \\ \hline
		\cite{mhala2025enhancing} & Data augmentation, L2 regularization, transfer learning & Potato disease dataset & 81.31\% accuracy\\ \hline
		\cite{singh2023zero} & Zero-shot transfer learning & Tomato & -\\ \hline
		\cite{sathya2025enhancing} & Conditional GANs & multiple plant & 98.78\% accuracy \\ \hline
		\cite{wu2023classification} & ConvNeXt with attention and random masking & Soybean leaf disease dataset & 85.42\% accuracy \\ \hline
		\cite{karthikeyan2023deep} & DenseNet-based CBIR (DTLDN-CBIRA) & multiple plant & 100\% accuracy, 81.90\% recall, 89.90\% F1-score \\ \hline
		\cite{haikal2024comprehensive} & Mixed augmentation (Cutout, CutMix, MixUp, FMix) & Rice leaf disease dataset & 91.51\% (ResNet34), 99.75\% (MobileNetV3) \\ \hline
		\cite{chen2024improved} & Enhanced FasterViT (CNN + ViT) & citrus  & - \\ \hline
	\end{tabular}
\end{table}

\subsection{Review of Olive Disease Classification Techniques}

Deep learning has revolutionized artificial intelligence, offering powerful solutions for accurately and rapidly classifying plant diseases, including those affecting olive trees. This section presents a detailed overview of recent advances in olive disease classification.

\cite{cruz2017x} created a deep learning framework that uses computer vision to identify OQDS, with a true positive rate of 98.60 ± 1.47\%.

\cite{uuguz2021classification} carried out tests comparing models with and without data augmentation using hand-gathered photos of olive leaves. The accuracy of augmentation-trained models was 95\%.

\cite{alshammari2022optimal} used a deep learning pipeline with a genetic algorithm for best results. The accuracy of their model was 98\% in binary classification and 96\% in multi-class tasks.

\cite{alshammari2022olive} suggested a hybrid architecture that combines CNNs and Vision Transformers, achieving 97\% accuracy for binary classification and roughly 96\% for multi-class issues.

\cite{ksibi2022mobires} presented MobiRes-Net, a ResNet50/MobileNet hybrid that was trained using 5,400 photos obtained by UAVs. The accuracy of the model was 97.08\%.

\cite{alshammari2023identification} created an enhanced ANN for feature selection and classification by utilizing the Whale Optimization Algorithm (WOA). All evaluation parameters, including F1-score, precision, recall, and accuracy, showed gains with this approach.

\cite{alsaeedi2023dynamic} offered a five-step methodology that includes deep learning classification, feature extraction, clustering, and picture enhancement. Their model achieved 98\%  accuracy with more than 3,200 photos, surpassing both VGG16 (96\% ) and AlexNet (95\%).

\cite{mahmood2024olive} suggested a technique that combines pre-trained networks with wavelet transforms to obtain highly discriminative features for precise illness identification.

\cite{bruno2023efficient} presented a two-model ensemble technique utilizing EfficientNet-b0. The classification of olive diseases was made more accurate and resilient by this adaptive method.

A summary of these approaches is provided in \autoref{table.2}. While the reported models demonstrate high performance, they also highlight persistent challenges such as overfitting, limited dataset diversity, and computational constraints.

\begin{table}[!ht]
	\centering
	\caption{Summary of Studies on Olive Disease Classification}
	\label{table.2}
	\begin{tabular}{p{2cm} p{4cm} p{4cm} p{2cm} p{2cm}}
		\hline
		\textbf{Study} & \textbf{Methods} & \textbf{Dataset} & \textbf{Accuracy} & \textbf{Data Generation Type} \\ \hline
		\cite{cruz2017x} & Computer vision-based deep learning & Olive leaf (1 disease + healthy) & 98.60 ± 1.47\% & None \\ \hline
		\cite{uuguz2021classification} & Deep learning (with augmentation) & Olive leaf (2 diseases + healthy) & 95\% & Traditional Augmentation \\ \hline
		\cite{alshammari2022optimal} & Deep learning + genetic algorithm & Olive leaf (2 diseases + healthy) & 96\% (multi-class), 98\% (binary) & Traditional Augmentation \\ \hline
		\cite{alshammari2022olive} & CNN + Vision Transformer & Olive leaf (2 diseases + healthy) & 97\% (binary), ~96\% (multi-class) & Traditional Augmentation  \\ \hline
		\cite{ksibi2022mobires} & MobiRes-Net (ResNet50 + MobileNet) & UAV-acquired images (5,400 samples) & 97.08\% & Traditional Augmentation \\ \hline
		\cite{alshammari2023identification} & Optimized ANN + WOA & Olive leaf (2 diseases + healthy) & 98\% & Traditional Augmentation  \\ \hline
		\cite{alsaeedi2023dynamic} & Image enhancement + DNN & 3,200 images (2 diseases + healthy) & 98\% & Traditional Augmentation \\ \hline
		\cite{mahmood2024olive} & Wavelet transform + pre-trained DL models & Olive leaf (2 diseases + healthy) & 98\% & Traditional Augmentation  \\ \hline
		\cite{bruno2023efficient} & Ensemble of EfficientNet-b0 models & Olive leaf (2 diseases + healthy) & 98\% & Traditional Augmentation  \\ \hline
		\textbf{Proposed Work} & Traditional and advanced data generation (CutMix, MixUp, DCGAN) across 3 distribution scenarios (S1, S2, S3) & Olive leaf (6 diseases + healthy) dataset (6170 images) & 93.48\% & Traditional and (CutMix, MixUp, DCGAN) \\ \hline
	\end{tabular}
\end{table}

As shown in \autoref{table.2}, the majority of previous studies rely on classical data generation techniques, or even no explicit generation at all. These approaches often achieve high accuracy, but on relatively homogeneous datasets. However, differences in accuracy can also be attributed to the diversity of data used in different studies. For example, some studies used smaller or specific datasets, while others worked with larger and more diverse data. The proposed work combines traditional and advanced data generation methods-such as CutMix, MixUp, and DCGAN-to simulate more varied distribution scenarios. Although the accuracy obtained (93.48\%) is slightly lower than that of some conventional methods, this compromise allows greater robustness and generalizability in the face of real data variations. The difference in accuracy reflects not only the choice of data generation techniques but also the inherent differences in the datasets used, which probably led to distinct model behavior and performance.

\subsection{Research Gap Analysis and Contributions}

Despite the promising advances discussed, most existing studies either focus on a single augmentation strategy or fail to explore the impact of data distribution scenarios on classification performance. Additionally, few works systematically integrate both classical and generative generation techniques while evaluating their effects across multiple imbalance contexts. To bridge this gap, the proposed work introduces a robust and comprehensive data generation strategy combining traditional augmentation, CutMix, MixUp, and DCGAN, evaluated under three distinct distribution scenarios (S1, S2, S3) for olive disease classification. This approach aims to enhance both the generalization and the robustness of deep learning models under challenging real-world constraints.

\subsection{Research Contributions}

This study suggests a data generation technique to enhance the classification of diseases affecting olive trees. The following is a summary of this work's primary contributions:

\begin{enumerate}
	\item \textbf{Hybrid Data Generation Framework:} We introduce a hybrid strategy that combines traditional augmentation techniques with MixUp, CutMix, and  DCGAN. This method improves classification performance, especially for underrepresented classes, by increasing the dataset's diversity and representativeness.
	
	\item \textbf{Mitigating Class Imbalance:} The proposed framework explicitly addresses the issue of class imbalance by generating additional samples for minority classes and adjusting the overall data distribution. This results in a more balanced training dataset, reduces model bias, and improves overall accuracy.
	
	\item \textbf{Dataset Expansion Strategy:} Beyond combining traditional augmentations, CutMix, MixUp, and DCGANs, we evaluate the framework under three distinct distribution scenarios (S1, S2, S3) for olive disease classification. This strategy enhances the model's generalizability and reduces the risk of overfitting, especially in data-scarce contexts.
	
	\item \textbf{Comprehensive Performance Evaluation:} We carefully investigate the proposed data generation techniques to ascertain their impact on the classification accuracy model and robustness in identifying olive tree diseases.
\end{enumerate}

\subsection{Problem Description}

The impact of data generation methods on the categorization of diseases affecting olive trees is investigated in this study. We combine traditional techniques (such as geometric and color transformations) with cutting-edge strategies like Mixup, CutMix, and DCGAN to address problems of data scarcity and class imbalance. Improving model generalization and accuracy, particularly for underrepresented classes, is the goal.

Key challenges include:

\begin{itemize}
	\item \textbf{Class Imbalance:} Certain disease classes are under-sampled, leading to biased models.
	\item \textbf{Overfitting:} Too much or unwarranted augmentation can add artefacts, heightening the danger of overlearning.
	\item \textbf{Selecting Effective Augmentation:} To enhance performance without sacrificing robustness, one needs to choose the right methods for a specific dataset and task.
\end{itemize}
This study compares and evaluates different augmentation techniques to improve generalisation and avoid over-fitting in a bid to improve plant disease detection models in low-data scenarios.

\section{Solution Approach}\label{sec3}

\begin{figure*}[!ht]
	\centering
	\includegraphics[width=\linewidth]{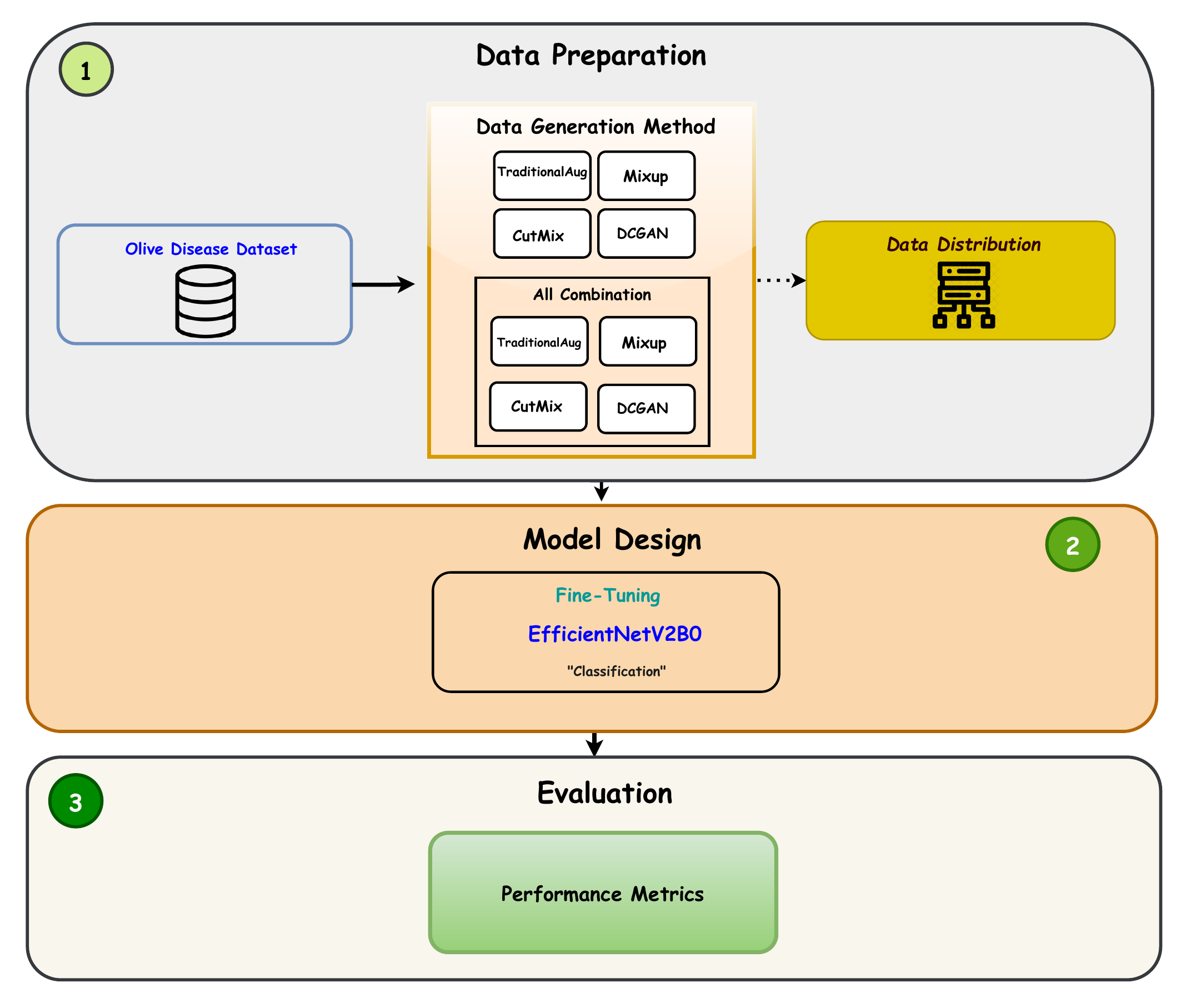}
	\caption{The proposed block work.}
	\label{Fig.1}
\end{figure*}

The primary phases of the proposed methodology, data preparation, model development, and performance evaluation, are explained in this section. Two subsets of the data were initially prepared: Dataset A  for training and Dataset B for evaluating generalization as a new test set. To obtain a more balanced and diverse training dataset, certain data generation techniques, such as synthesis by DCGAN, MixUp, CutMix, and traditional transformations, were carried out on Dataset A. Subsequently, to ensure an equal comparison across the different experimental conditions (S1, S2, S3), the original and generated samples were split into training and validation sets. These datasets were used for training and fine-tuning a deep learning architecture, i.e., EfficientNetV2B0, to develop the model. Finally, the performance of the model was assessed on Dataset B, which was untouched during training, to get an unbiased estimation of each scenario's performance in olive disease classification. The entire simulation workflow is shown in \autoref{Fig.1}.

\subsection{Data Pereparation  }

\subsubsection{Data Collection}

\begin{figure*}[!ht]
	
	\centering
	\includegraphics[width=\linewidth]{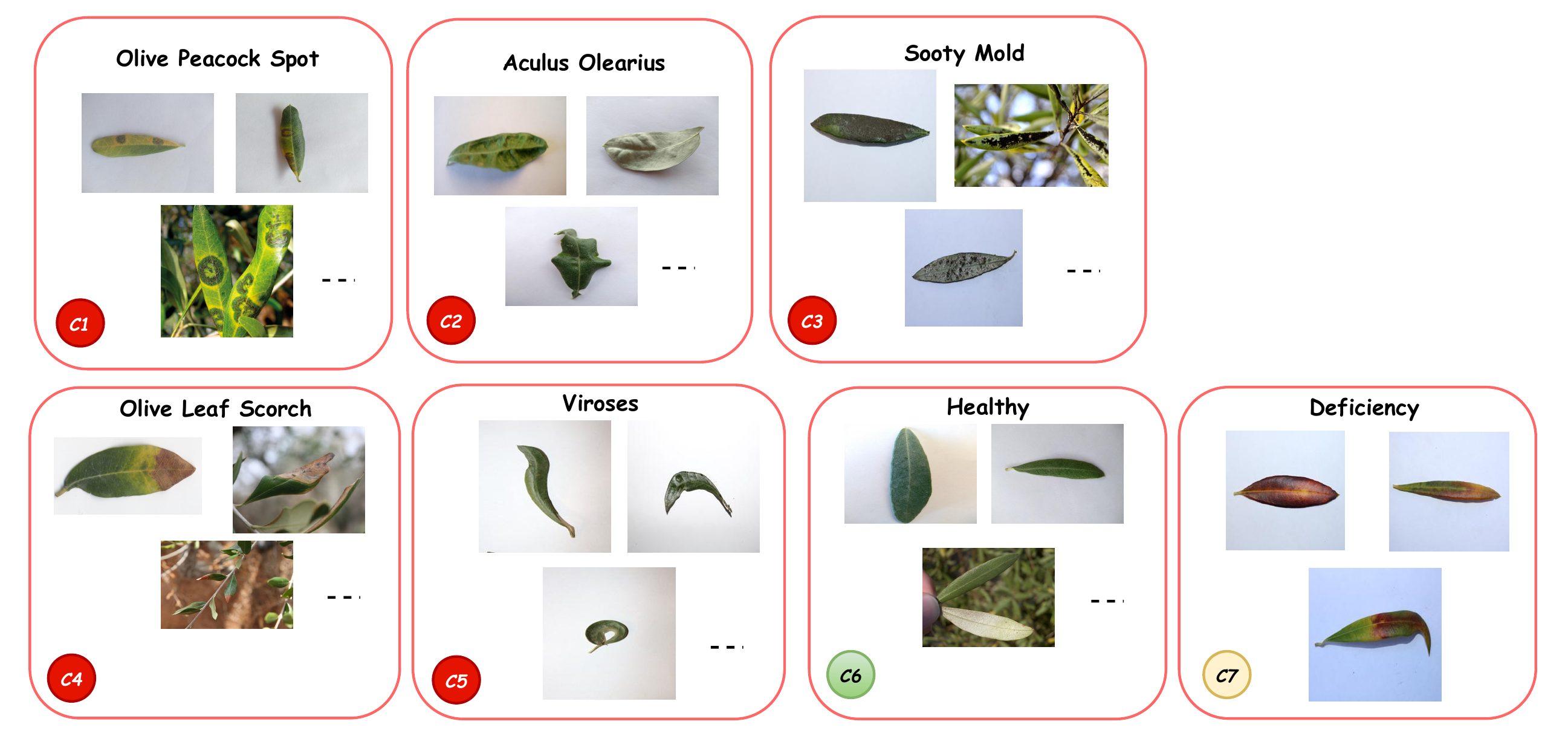}
	\caption{Example images depicting various categories of diseases affecting olive leaves.}
	\label{fig2}
\end{figure*}

In this study, images of olive leaves were collected from several sources, including datasets from previous research. Photographs taken on Moroccan olive farms between 2020 and 2024 are included in \footnote{\url{https://github.com/sinanuguz}}, \footnote{\url{https://github.com/DrAlbertCruz/OQDS-GY-Dataset}}, \footnote{\url{https://github.com/ichaparroc/OliveLeafDiseaseDatasetTacna}}, and \footnote{\url{https://www.kaggle.com/datasets/serhathoca/zeytin}}. A single olive leaf, represented by each image, can be categorized as either healthy or afflicted by one of the following disorders. The entire dataset statistics are summarized in \autoref{table.1}, and representative photos from each category are shown in \autoref{fig2}.

\begin{table*}[!ht]
	\centering
	\caption{Image count per class in the dataset}
	\label{table.1}
	\begin{tabular}{@{}p{2cm}p{3cm}p{6cm}p{6cm}@{}}
		\toprule
		\textbf{Plant} & \textbf{Class} & \textbf{DataSet A} & \textbf{DataSet B (Test)} \\
		\midrule
		\textbf{Olive} & Healthy & 1292 & 471 \\
		& Aculus Olearius & 712 & 384 \\
		& Olive Peacock Spot & 1716 & 703 \\
		& Olive Leaf Scorch & 68 & 43 \\
		& Sooty mold & 205 & 56 \\
		& Deficiency & 206 & 56 \\
		& Virosis & 206 & 52 \\
		\addlinespace
		\midrule
		\bottomrule
	\end{tabular}
\end{table*}

\subsection{Data Generation}\label{subsubsec2}
Data generation techniques are becoming more and more important in modern deep learning applications, both to improve model generalization on heterogeneous and imbalanced datasets and to avoid overfitting. This work uses a variety of data creation techniques, including as classical augmentation, CutMix, MixUp, DCGAN, and a thorough mix of these approaches, to improve the performance of the olive illness classification model.

\textbf{Traditional Augmentation:} Several standard data augmentation methods, such as rotation, flipping vertically and horizontally, and zooming, were used to enlarge dataset A. The training data became more varied and variable as a result of this procedure, which produced a dataset that was around four times larger than the original.

\textbf{CutMix:} CutMix \cite{yun2019cutmix} is a regularization method that creates fresh training samples by merging the labels of various images proportionately and clipping and pasting patches between them. After pasting a patch from \( x_B \) onto \( x_A \) given two input pictures \( x_A \) and \( x_B \), the new image \( \tilde{x} \) and label \( \tilde{y} \) are defined as follows:

\begin{equation}
	\tilde{x} = M \odot x_A + (1 - M) \odot x_B \quad \text{(Eq. 1)}
\end{equation}

\begin{equation}
	\tilde{y} = \lambda y_A + (1 - \lambda) y_B \quad \text{(Eq. 2)}
\end{equation}

where \( M \) is a binary mask indicating the patched region, \( \odot \) denotes element-wise multiplication, and \( \lambda \) is sampled from a Beta distribution.

\textbf{MixUp:}   \cite{zhang2017mixup} generates new training samples by linearly interpolating two random images \( x_i \) and \( x_j \), and their labels \( y_i \) and \( y_j \), using a mixing parameter \( \lambda \in [0,1] \):

\begin{equation}
	\tilde{x} = \lambda x_i + (1 - \lambda) x_j \quad \text{(Eq. 3)}
\end{equation}

\begin{equation}
	\tilde{y} = \lambda y_i + (1 - \lambda) y_j \quad \text{(Eq. 4)}
\end{equation}

This method encourages the model to behave linearly in-between training examples, enhancing generalization and reducing overfitting.

\textbf{DCGAN:} In this study, we utilized Deep Convolutional Generative Adversarial Networks (DCGANs)  \cite{radford2015unsupervised} to generate synthetic images of olive diseases. A DCGAN consists of a generator \( G \) and a discriminator \( D \) trained in a minimax game:

\begin{equation}
	\min_G \max_D \; \mathbb{E}_{x \sim p_{\text{data}}(x)} [\log D(x)] + \mathbb{E}_{z \sim p_z(z)} [\log(1 - D(G(z)))] \quad \text{(Eq. 5)}
\end{equation}

where \( z \) is a random noise vector. Through iterative training, \( G \) learns to generate realistic images, while \( D \) learns to distinguish real from fake samples.

\textbf{Frechet Inception Distance (FID):} To measure the quality of generated images, we used the FID metric \cite{jayasumana2024rethinking} which computes the distance between the feature distributions of real and synthetic data:

\begin{equation}
	\text{FID} = \left\| \mu_r - \mu_g \right\|^2 + \text{Tr}\left( \Sigma_r + \Sigma_g - 2(\Sigma_r \Sigma_g)^{1/2} \right) \quad \text{(Eq. 6)}
\end{equation}

where \( (\mu_r, \Sigma_r) \) and \( (\mu_g, \Sigma_g) \) are the mean and covariance of features extracted from real and generated images using an Inception network.

\textbf{In summary}, all of these approaches were employed to generate new data samples, which were subsequently distributed over the originally collected dataset to improve class balance and enhance the model’s performance.

\subsection{Data Distribution}\label{subsubsec3}

After collecting olive leaf disease samples and expanding the dataset using data augmentation techniques, we designed a clear and structured strategy to distribute the data. We defined three main experimental setups, each with a different way of organizing the data. For each setup, we created a specific distribution plan to better understand how different data arrangements influence the model’s performance.

\begin{enumerate}
	\item \textbf{S1: Balancing class distributions in the original dataset:} In this case, new data was used to balance the number of samples between classes in the original dataset. This process was unavoidable for reducing model bias against majority classes and enhancing classification fairness.
	
	\item \textbf{S2: Increasing the sample size of each class:} In this case, the intention was to enhance the dataset by adding more samples per class using the created data. Enhancing the learning and generalization of the model was the intention of this improvement.
	
	\item \textbf{S3: Balancing class distributions while expanding the dataset size:} In this instance,  class balancing goals and dataset expansion goals were combined. Not only did it increase the overall amount of data, but also gave a balanced view of all classes, which is essential in order to get the highest possible accuracy and potency of the model.
\end{enumerate}

\subsection{Model design}\label{subsubsec4}

\begin{figure*}[!ht]
	\centering
	\includegraphics[width=\textwidth]{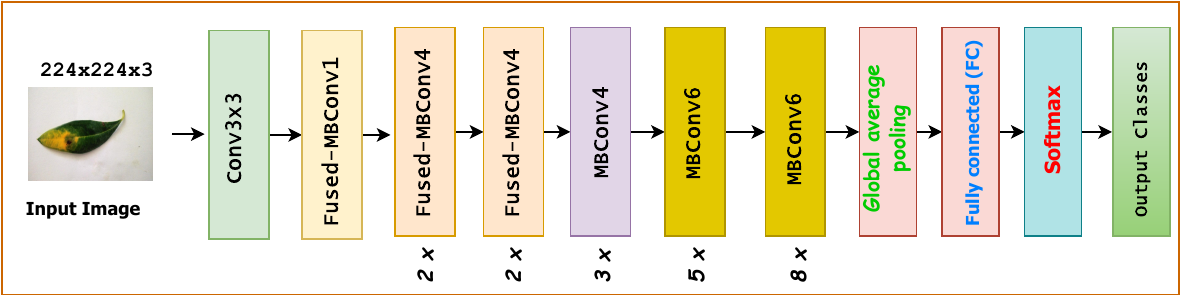}
	\caption{The architecture proposed for the classification.}\label{fig7}
\end{figure*}

This study uses leaf imagery to identify diseases in olive trees using the pre-trained EfficientNetV2B0 model. Acknowledged for its effectiveness and precision in object recognition tasks, this architecture gains from previous training on large datasets like ImageNet and ImageNet21k \cite{pmlr}. We modified the model to fit our classification task by substituting a custom classifier consisting of four fully connected layers, each followed by a Dropout layer, and ending with a softmax activation (see \autoref{fig7}). Although several deep learning models were tested during the experiment, EfficientNetV2B0 continuously produced better results. Existing research demonstrates that its selection was justified by its balance of high predictive performance, fast convergence, and computational efficiency.

\subsection{Perfermence Evalution }\label{subsubsec5}

Several evaluation metrics were used in this study to measure the performance of the olive disease classification model. The indicators used include accuracy, F1 score, precision, recall, Matthews Correlation Coefficient (MCC), and geometric mean (G-mean) \cite{chicco2020advantages,de2024performance}. For a more thorough examination, the Non-Parametric Scott-Knott ESD (NPSK) test \citep{tantithamthavorn2016empirical, tantithamthavorn2018impact} was also used. The main metric was accuracy, which was validated using 5-fold cross-validation (k = 5). The Scott-Knott ESD test uses hierarchical clustering to group methods and determine the best distribution strategy. It is implemented using the ScottKnottESD package in R.

\subsection{Computational Resources and Time Requirements}\label{Ex}

The experimental configuration used to implement our proposed method and evaluate its computational cost is presented in \autoref{tab:c}.

\begin{tcolorbox}[colback=blue!5!white, colframe=blue!75!black]
	\begin{itemize}
		\item \textbf{Operating System:} Windows 10
		\item \textbf{Software:} TensorFlow, configured to run on the above operating system
		\item \textbf{RAM:} 46 GB
		\item \textbf{GPU:} NVIDIA GeForce RTX 2060 with 12 GB of dedicated RAM
		\item \textbf{CPU:} Intel(R) Xeon(R) E5-2630 v3, 2.40 GHz
	\end{itemize}
\end{tcolorbox}

\begin{table*}[h!]
	\caption{Computational Costs Associated with the Proposed Methodology} \label{tab:c}
	\begin{tabular}{@{}p{8cm}p{6cm}@{}}
		\toprule
		\textbf{Process} & \textbf{Approximate Duration} \\
		\midrule
		\textbf{ Data Generation using DCGAN (Generated Data)} & ~3 months \\
		\addlinespace
		\textbf{Traditional Data Augmentation (e.g., Flip, Rotation)} & ~1 hour \\
		\addlinespace
		\textbf{Data generation using MixUp and CutMix} & ~1 hour \\
		\addlinespace
		\textbf{Selection of High-Quality (Generated Data) using Frechet Inception Distance (FID)} & ~2 months \\
		\addlinespace
		\textbf{Training of Classification Models (EfficientNetV2B0)} & ~30 minutes per experiment (see \autoref{tab:metrics}) \\
		\bottomrule
	\end{tabular}
\end{table*}

\section{Results Analysis}\label{sec4}

\autoref{tab:metrics} presents the classification accuracy obtained with different data generation techniques in three experimental scenarios (S1, S2, S3), compared to a baseline model without augmentation (accuracy = 90.99\%).

\textbf{\textbf{S1}: Balanced classes }

In this scenario, all the methods outperformed the baseline. The most significant improvement was seen with the full combination of augmentations (Traditional + MixUp + CutMix + DCGAN), achieving an accuracy of 93.09\%, an improvement of +2.10 points on the baseline. CutMix alone also improved significantly by +0.91 points, followed by DCGAN (+1.25) and MixUp (+0.51). This shows that data augmentation is particularly beneficial in balanced contexts.

\textbf{\textbf{S2}: Increased class volume (imbalanced)}

The results were more mixed. Only the MixUp method significantly improved performance with +0.74 points. The full combination of augmentations maintains a slight improvement (+0.34), while the traditional (-0.23) and CutMix (-0.28) methods underperform. This indicates that certain augmentations can accentuate the imbalance, whereas MixUp, with its linear interpolation, favours better regularisation.

\textbf{\textbf{S3}: Balanced classes + increased volume}

In this mixed scenario, CutMix is the best performing method with +1.30 points, demonstrating its effectiveness with larger datasets. DCGAN and MixUp show moderate gains (+0.40 and +0.11, respectively), while the full combination brings no net gain (+0.00), suggesting saturation or redundancy of effects.

\begin{table}[!ht]
	\centering
	\caption{Test accuracy on the olive disease dataset using various data generation techniques.}
	\label{tab:metrics}
	\begin{tabular}{lccccccccccc}
		\toprule
		& \multicolumn{1}{c}{No Aug.} & \multicolumn{2}{c}{Traditional Aug.} & \multicolumn{2}{c}{MixUp} & \multicolumn{2}{c}{CutMix} & \multicolumn{2}{c}{DCGAN} & \multicolumn{2}{c}{All Comb.} \\
		\cmidrule(r){2-2} \cmidrule(r){3-4} \cmidrule(r){5-6} \cmidrule(r){7-8} \cmidrule(r){9-10} \cmidrule(r){11-12}
		& Acc. & Acc. & Diff. & Acc. & Diff. & Acc. & Diff. & Acc. & Diff. & Acc. & Diff. \\
		\midrule
		S1 & --    & 91.56 & +0.57 & 91.50 & +0.51 & 91.90 & +0.91 & 92.24 & +1.25 & \textbf{93.09} & +2.10 \\  
		S2 & --    & 90.76 & -0.23 & 91.73 & +0.74 & 90.71 & -0.28 & 91.33 & +0.34 & \textbf{93.48} & +2.49 \\  
		S3 & --    & 90.76 & -0.23 & 91.10 & +0.11 & \textbf{92.29} & +1.30 & 91.39 & +0.40 & 90.93 & -0.06 \\  
		\midrule
		Baseline & 90.99 & --    & --    & --    & --    & --    & --    & --    & --    & --    & --    \\  
		\bottomrule
	\end{tabular}

	\footnotesize{
		\textbf{S1}: Balanced classes \\
		\textbf{S2}: Increased class volume (imbalanced) \\
		\textbf{S3}: Balanced classes + increased volume \\
	}
\end{table}

\autoref{tab:wildmetrics} presents the evaluation scores (Accuracy, Recall, F1-score, G-mean, and MCC) obtained on the olive disease dataset for different scenarios (S1, S2, S3) and generation methods.

In scenario S1 (balanced classes), the best overall performance is observed with the combined approach (all augmentations), which achieves an F1-score of 91.89\%, a G-mean of 90.37\% and an MCC of 90.15\% , outperforming the baseline (F1-score = 89.63\%). Using DCGAN alone also shows a significant improvement with an F1-score of 93.14\% , suggesting better generalisation through the generation of new synthetic images.

In scenario S2 (volume increase without balancing), the results are even more significant. The combined approach obtains the highest scores on almost all indicators, reaching 95.99\% for accuracy, 93.32\% for F1-score and 91.04\% for G-mean, indicating that the joint increase in data and its diversity are particularly beneficial in cases of class imbalance. Data generation with DCGAN alone also performed very well (F1-score = 93.35\%), demonstrating the ability of this generator to fill gaps in under-represented classes.

For scenario S3 (balanced classes with more volume), the trend is more nuanced. Although CutMix provides the best results in terms of F1-score (92.28\%) and MCC (89.48\%), the combined approach does not always outperform the isolated methods. This could indicate redundancy or a saturation effect when several techniques are used simultaneously on a game that is already sufficiently rich and balanced.
In summary, the results highlight that generation techniques, particularly DCGAN and advanced combinations (MixUp, CutMix), significantly improve the model's robustness, particularly in scenarios of imbalance or low data volume.

\begin{table}[!ht]
	\centering
	\caption{Evaluation scores on the test dataset (\%).}
	\label{tab:wildmetrics}
	\begin{tabular}{lcccccc}
		\toprule
		\textbf{} & \textbf{Data Generation} & \textbf{Precision} & \textbf{Recall} & \textbf{F1-Score} & \textbf{G-mean} & \textbf{MCC} \\
		\midrule
		Baseline & None & 92.14 & 88.04 & 89.63 & 87.66 & 87.53 \\
		\midrule
		\multirow{5}{*}{S1}
		& Traditional Aug. & 92.38 & 88.67 & 89.89 & 87.99 & 88.35 \\
		& Mixup & 92.05 & 91.50 & 91.48 & 89.19 & 88.46 \\
		& Cutmix & 92.17 & 91.90 & 91.91 & 89.25 & 88.77 \\
		& DCGAN & 93,66   & 93,09  &  93,14 & 90,55 & 90,55 \\
		& All Combination & 93.89 & 90.70 & 91.89 & 90.37 & 90.15 \\
		\midrule
		\multirow{5}{*}{S2}
		& Traditional Aug. & 90.92 & 90.76 & 90.79 & 89.06 & 87.12 \\
		& Mixup & 91.88 & 91.73 & 91.71 & 89.23 & 88.52 \\
		& Cutmix & 91.38 & 90.71 & 90.66 & 88.28 & 87.44 \\
		& DCGAN & 94.00 & 93.48 & 93.35 & 90.58 & 91.12 \\
		& All Combination & 95.99 & 91.48 & 93.32 & 91.04 & 90.89 \\
		\midrule
		\multirow{5}{*}{S3}
		& Traditional Aug. & 91.04 & 90.76 & 90.76 & 88.03 & 87.20 \\
		& Mixup & 91.40 & 91.10 & 91.11 & 87.66 & 87.64 \\
		& Cutmix & 92.84 & 92.29 & 92.28 & 88.63 & 89.48 \\
		& DCGAN & 91.39 & 91.39 & 91.43 & 88.61 & 88.61 \\
		& All Combination & 91.21 & 90.93 & 90.94 & 88.14 & 87.39 \\
		\bottomrule
	\end{tabular}
\end{table}

\begin{figure*}
	\centering
	\includegraphics[width=\textwidth]{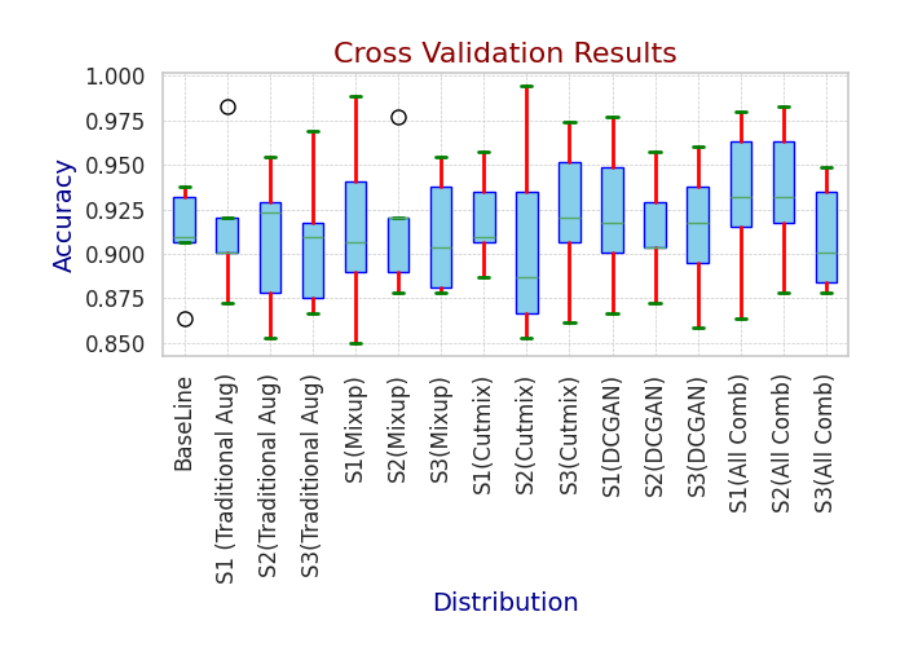}
	\caption{Cross-validation Results.}
	\label{fig8}
\end{figure*}

\autoref{fig8} presents the results of a cross-validation procedure applied to evaluate the ranking of the training data sets used in this study. The figure visualizes the comparative performance of the different training datasets through a ranking system, highlighting the models' responses when trained on these datasets. This ranking helps identify which dataset configurations have the most significant impact on model performance, providing valuable insights into data selection and distribution strategies.

\begin{figure*}
	\centering
	\includegraphics[width=\textwidth]{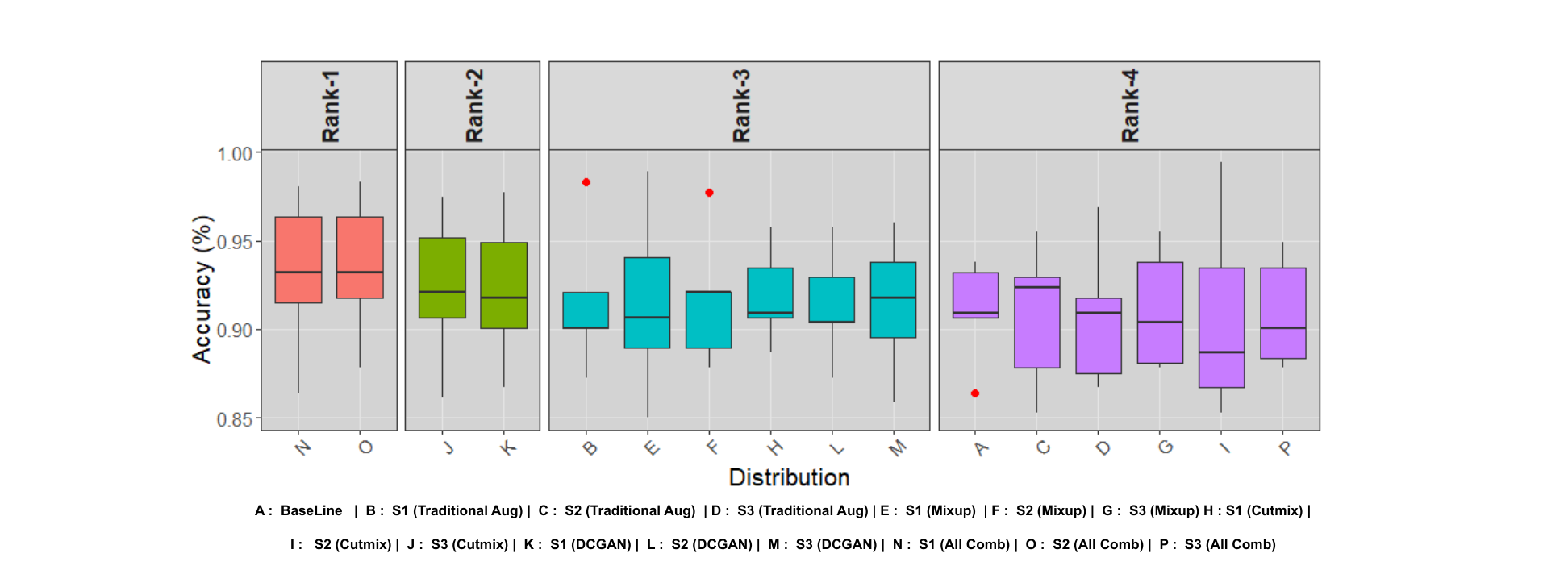}
	\caption{Ranking of Training Data Distributions Utilized in this Study for Model Training.}
	\label{fig9}
\end{figure*}

\autoref{fig9} illustrates another key aspect of model evaluation, focusing on the ranking of training data sets as assessed during the study. The figure compares the impact of various datasets on model performance, with each set ranked based on its efficacy in improving the model's accuracy. By analyzing these rankings, the study aims to identify the most effective data subsets for optimizing model training.

\section{Discussion}\label{sec5}

The experimental results obtained using different scenarios and data generation techniques reveal several key insights into improving the performance of classification models, particularly in the context of olive disease detection.
Initially, traditionel augmentation methods such as rotation and inversion offer a modest yet consistent enhancement over baseline performance, highlighting their value in typical scenarios. Nevertheless, their standalone effectiveness is constrained, especially in cases involving class imbalance or substantial inter-class resemblance. These techniques introduce only limited variation to the original dataset, contributing minimal new information to the model, which accounts for the relatively minor performance gains observed.

The integration of advanced augmentation methods, such as MixUp and CutMix, shows more competitive results, particularly in scenarios where the classes are balanced (S1) or enriched (S3). These techniques, based on the creation of interpolated or hybrid images, facilitate the learning of stronger and more generalizable representations by the model, which translates into better generalization over the test set. However, these methods have certain limitations: in some cases, they can introduce artificial samples that are not very informative for learning. Their effectiveness, therefore, depends heavily on the application scenario and their judicious use within a justified framework.

The generation of synthetic images using DCGAN has proved remarkably effective, particularly in scenarios involving imbalance (S2). By enriching the minority classes with realistic artificial samples, DCGAN reduces the impact of the bias induced by the initial distribution of the data. This ability is particularly highlighted by the high F1-score and MCC in S2, where the synthetic data compensates for the lack of diversity. The use of DCGAN has a positive influence on the performance of the classification model, as it makes it possible to generate diverse samples from the original data. However, this approach also raises ethical constraints linked to the quality and validity of the data generated. This is why, in this study, we used the Fréchet distance to select only high-quality samples, so as not to disrupt the model learning process.

The combined approach, combining all the techniques (Traditional, MixUp, CutMix, DCGAN), generally gives the best overall performance, particularly in scenarios S1 and S2. This confirms that diversification of augmentation strategies can act in a complementary way, increasing the quantity, variability, and representativeness of the training data.

However, in scenario S3 (data already balanced and voluminous), the effect of certain combinations seems to diminish. This phenomenon can be explained by model saturation, where the excessive addition of modified or synthetic samples no longer provides useful information and may even introduce noise.
In summary, this study underscores the significance of tailoring data generation strategies to the unique attributes of a given dataset. The thoughtful application of advanced techniques, especially DCGAN and well-planned combinations, proves to be a crucial factor in enhancing computer vision performance, both in agricultural contexts and broader domains.

\subsection{Sensitivity Analysis}

A sensitivity analysis was carried out by comparing the relative impact of each method of increase on the different scenarios:

S1 (balanced classes): All augmentation methods showed significant gains, with a more marked sensitivity for CutMix (+0.91) and DCGAN (+1.25), indicating their relevance even on well-structured datasets.

S2 (augmented volume): The performance of traditional augmentations and CutMix is less sensitive or even negatively impacted (e.g., CutMix at -0.28), while MixUp and DCGAN remain robust, with a positive of +0.74 and +0.34, respectively. This shows that the robustness of the techniques strongly depends on the initial imbalance of the data.

S3 (balanced + voluminous): The gains are more moderate, even zero for some combinations. This suggests performance saturation, as the model has already been exposed to sufficient diversity. However, CutMix retains an advantage (+1.30), indicating that it can still enrich the representations, even in well-fed cases.

In summary, the sensitivity of performance to augmentation methods depends closely on the characteristics of the dataset: volume, balance, and intra-class diversity. DCGAN and CutMix stand out as the most efficient and stable, while classical methods are to be preferred in simple cases or for low-variance tasks.

\subsection{Ethical Considerations in Data Generation Techniques}

Use of advanced generation methods like DCGAN, MixUp, and CutMix is of serious ethical concern, especially in sensitive areas like agricultural disease detection.

To guarantee ethical integrity and minimize any potential biases stemming from our approach to generating data, the following actions were taken:
\begin{itemize}
	\item \textbf{Ethical Practices:} We promote ethical handling of data during the process of generating and collecting the data. We are committed to preventing artificial bias or unfair results, especially by systematically visualizing the generated data to ensure that the resulting samples are unbiased and of high quality.
	\item \textbf{Bias Mitigation Strategies:} To reduce potential model bias, we carefully select data sources and actively monitor class distributions for imbalances. This ensures that the generated data supports real samples in a representative and balanced manner.
	\item \textbf{Transparency and Fairness:} Transparency in the generation and selection of synthetic data is a key objective. For images generated by DCGAN, we employ the Fréchet Inception Distance (FID) to evaluate and retain the highest-quality samples.
\end{itemize}

By integrating these ethical considerations into our solution, we aim to foster fair, transparent, and replicable AI practices in precision agriculture.

\section{Conclusions, limitations, and future research}\label{sec6}

\subsection{Findings}

This paper introduces a new method for olive tree disease classification by combining traditional data augmentation methods with state-of-the-art approaches such as MixUp, CutMix, and Deep Convolutional Generative Adversarial Networks (DCGAN). Experiments on a multi-class dataset with seven classes, namely disease-free olive leaves and other disease states, indicate that the proposed data generation strategy in this paper significantly enhances model accuracy and robustness.

The results confirm that data generation plays a crucial role in improving classification performance, particularly when used in well-defined scenarios rather than applied randomly. The proposed approach achieved a performance improvement of 2.48\%  compared to the baseline (no augmentation).

\subsection{Future Research}
Future research could explore the following avenues to build upon and extend the findings of this study:
\begin{itemize}
	\item \textbf{Exploration of Additional Generation Techniques:} To better address issues with intra-class variability and inter-class similarity, future research should look at more sophisticated data generation methods.
	\item \textbf{Real-World Deployment and Validation:} Applying the proposed method in real agricultural settings would allow for the evaluation of its practical viability, robustness, and adaptability under real-world conditions.
\end{itemize}

\subsection{Limitations of Research}

Despite its promising results, this study has a few limitations:

\begin{itemize}
	\item \textbf{Data Dependency:} The quality and balance of the dataset have an impact on the model's performance. The effectiveness of the model's generalization may be hampered by incomplete or skewed data.
	\item \textbf{Computational Cost:} Techniques like DCGAN require significant resources, which can limit scalability.
	\item \textbf{Parameter Sensitivity:} The approach is sensitive to the volume of generated data and key model hyperparameters, requiring careful tuning for optimal results.
\end{itemize}


\bibliography{sn-bibliography}


\end{document}